\documentclass[sigconf]{acmart}

\usepackage[nolist]{acronym}
\usepackage{enumitem}
\usepackage{xcolor}
\usepackage{svg}
\usepackage[normalem]{ulem}
\usepackage{multirow}
\graphicspath{{./images/}}
\DeclareGraphicsExtensions{.svg,.pdf,.png} %
\svgpath{{./images/}} %
\usepackage{colortbl}

\AtBeginDocument{%
  \providecommand\BibTeX{{%
    \normalfont B\kern-0.5em{\scshape i\kern-0.25em b}\kern-0.8em\TeX}}}

\setcopyright{rightsretained}
\copyrightyear{2024}
\acmYear{2024}
\acmDOI{10.1145/3639476.3639760}

\acmConference[ICSE-NIER'24]{New Ideas and Emerging Results }{April 14--20, 2024}{Lisbon, Portugal}
\acmBooktitle{New Ideas and Emerging Results (ICSE-NIER'24), April 14--20, 2024, Lisbon, Portugal}
\acmPrice{15.00}
\acmISBN{979-8-4007-0500-7/24/04}

\begin{document}

\title{Designing Trustful Cooperation Ecosystems is Key to the New Space Exploration Era}

\author{Renan Lima Baima}
\orcid{0000-0002-6281-8153}
\affiliation{
  \institution{University of Luxembourg}
  \department{FINATRAX Research Group, SnT}
  \streetaddress{29, Avenue J.F Kennedy}
  \city{Kirchberg}
  \postcode{L-1855}
  \country{Luxembourg}}
\email{renan.limabaima@uni.lu}

\author{Chovet Lo{\"i}ck}
\orcid{0000-0003-4025-9095}
\affiliation{
  \institution{University of Luxembourg}
  \department{SpaceR Research Group, SnT}
  \streetaddress{29, Avenue J.F Kennedy}
  \city{Kirchberg}
  \postcode{L-1855}
  \country{Luxembourg}}
\email{loick.chovet@uni.lu}

\author{Johannes Sedlmeir}
\orcid{0000-0003-2631-8749}
\affiliation{
  \institution{University of Luxembourg}
  \department{FINATRAX Research Group, SnT}
  \streetaddress{29, Avenue J.F Kennedy}
  \city{Kirchberg}
  \postcode{L-1855}
  \country{Luxembourg}}
\email{johannes.sedlmeir@uni.lu}

\author{Miguel Angel Olivares Mendez}
\orcid{0000-0001-8824-3231}
\affiliation{
  \institution{University of Luxembourg}
  \department{SpaceR Research Group, SnT}
  \streetaddress{29, Avenue J.F Kennedy}
  \city{Kirchberg}
  \postcode{L-1855}
  \country{Luxembourg}}
\email{miguel.olivaresmendez@uni.lu}

\author{Gilbert Fridgen}
\orcid{0000-0001-7037-4807}
\affiliation{
  \institution{University of Luxembourg}
  \department{FINATRAX Research Group, SnT}
  \streetaddress{29, Avenue J.F Kennedy}
  \city{Kirchberg}
  \postcode{L-1855}
  \country{Luxembourg}}
\email{gilbert.fridgen@uni.lu}

\renewcommand{\shortauthors}{Baima et al.}

\begin{abstract}
  In the emerging space economy, autonomous robotic missions with specialized goals such as mapping and mining are gaining traction, with agencies and enterprises increasingly investing resources. \Ac{MRS} research has provided many approaches to establish control and communication layers to facilitate collaboration from a \emph{technical} perspective, such as granting more autonomy to heterogeneous robotic groups through auction-based interactions in mesh networks. However, stakeholders' competing \emph{economic} interests often prevent them from cooperating within a proprietary ecosystem.  Related work suggests that \ac{DLT} might serve as a mechanism for enterprises to coordinate workflows and trade services to explore space resources through a transparent, reliable, non-proprietary digital platform. We challenge this perspective by pointing to the core technical weaknesses of blockchains, in particular, increased energy consumption, low throughput, and full transparency through redundancy. Our objective is to advance the discussion in a direction where the benefits of \ac{DLT} from an economic perspective are weighted against the drawbacks from a technical perspective. We finally present a possible \ac{DLT}-driven heterogeneous \ac{MRS} for map exploration to study the opportunities for economic collaboration and competitiveness.%
\end{abstract}



\begin{acronym}
\acro{DLT}[DLT]{distributed ledger technology}
\acro{ISRU}[ISRU]{in situ resource utilization}
\acro{MRS}[MRS]{multirobot systems}
\acro{ESA}[ESA]{european space agency}
\end{acronym}

\received{4 September 2023}
\received[revised]{22 November 2023}
\received[accepted]{21 December 2023}

\maketitle

\section{Introduction}
\label{sec:introduction}

The development of the new space economy was bootstrapped thanks to policies and agreements by governments and international organizations~\cite{al-akharSpaceScienceTechnology2019}. The industry's transition from a centralized model to a decentralized free market system~\cite{orlovaPresentFutureSpace2020} has increased the number of companies and startups in the new space economy, with 80\% of the revenue generated in the last 16 years coming from private investments~\cite{connGlobalSpaceEconomy2021}. Major space agencies, including NASA~\cite{nasaSpaceTechnologyRoadmaps2016}, have outlined goals for \ac{ISRU}, such as destination reconnaissance, mapping, and resource acquisition. These plans aim to establish a long-term human presence on celestial bodies (e.g., Moon and Mars) through robotic lunar missions that utilize \ac{ISRU} technology~\cite{crawfordLunarResourcesReview2015} to perform tasks from mining to manufacturing solar panels.

\Ac{MRS} are groups of robots that collaborate to complete specific tasks or support each other based on individual capabilities. They can comprise both homogeneous and heterogeneous groups of robots and find applications in diverse fields such as precision farming, search and rescue, and space exploration~\cite{prettoBuildingAerialGroundRobotics2021}. The main advantage of using \ac{MRS} is that a team of geographically distributed robots can outperform single robots~\cite{deshpandeDecentralizedControlTeam2003}. However, its effectiveness hinges on robust, data-integrated coordination mechanisms that facilitate the exchange of norms, maintain team identity, and promote group confidence~\cite{strobelManagingByzantineRobots2018}.  To avoid the issue of information asymmetry in coordination mechanisms and the resulting ``market for lemons'' dilemma~\cite{akerlofMarketLemonsQuality1970}, it is crucial to minimize the knowledge gap between sellers and buyers. %
While financial incentives and sanctions can encourage cooperation and aid in meeting legal and regulatory requirements, managing resources and coordinating interactions among multiple organizations and countries can be challenging. In \emph{coopetitive} \ac{MRS}~\cite{ghobadiCoopetitiveKnowledgeSharing2011b}, various operators' economic incentives often compete, making it difficult to achieve common objectives and ensure participation. With over 60~countries involved in space activities~\cite{borowitzStrategicImplicationsProliferation2019}, many actors may be reluctant to join a centralized platform due to concerns about monopolistic or oligopolistic ownership, conflicts of interest, and treaty violations~\cite{leonMiningMeaningExamination2018}.

As organizations increasingly collaborate to achieve shared objectives~\cite{harrisonManagingPartneringExternal1996a}, deploying cross-border and cross-organization \ac{MRS} arguably becomes increasingly critical for future missions. However, the efficiency of \ac{ISRU} in such heterogeneous \ac{MRS} missions depends on effective communication and coordination among robots~\cite{martinezrocamoraMultirobotCooperationLunar2023}, such as the location and availability of critical resources, e.g., water and iron. The challenge of aligning the economic interests of competing stakeholders in a shared, non-proprietary digital space is expected to shape the new era of space exploration~\cite{orlovaPresentFutureSpace2020}. \Ac{DLT} has been proposed to facilitate collaboration among these entities, enabling autonomous, decentralized, peer-to-peer decision-making in \ac{MRS} for space missions~\cite{filippiBlockchainOuterSpace2021, ibrahimLiteratureReviewBlockchain2021}. In this regard, \ac{DLT} has shown promise in previous research ~\cite{khanBlockchainTechnologyApplications2019, filippiBlockchainOuterSpace2021}. In the context of space missions, it could allow robots to autonomously negotiate resource usage, such as shelters, without a central authority. However, it is essential to note that \ac{DLT} is not without technical limitations. Although it offers similar automation capabilities to centralized platforms through smart contracts, it is generally less efficient due to the inherent replication of transaction processing and storage~\cite{butijnBlockchainsSystematicMultivocal2020b, sedlmeir2020energy}. Consequently, deploying \ac{DLT} in \ac{MRS} and space exploration missions is particularly challenging~\cite{triantafyllouMethodologyApproachingIntegration2021a}, given the limited resources in such a remote environment. Additionally, the inherent transparency of blockchains complicates data access management, as information is either visible to all participants or none~\cite{sedlmeir2022transparency}.

This New Ideas and Emerging Results paper intentionally steers from presenting an exhaustive list of applications and challenges to showcase practical application development. It emphasizes exploring new ideas and the suitable needs of using \ac{DLT} for non-proprietary market-based coordination of \acp{MRS} in space. Our approach is not about pioneering a specific, feasible application in space exploration and multirobotic coordination. We revisited the market-based \ac{MRS} space mapping case presented by \citet{diasMarketBasedMultirobotCoordination2006c}, which we have already begun implementing to evaluate the next steps. Our primary contribution is investigating the economic benefits and assessing potential solutions’ technical challenges in space exploration requirements~\cite{lorenzHowFarFar2020, zlotMultirobotExplorationControlled2002a, garciaArchitectureDecentralizedCollaborative2018}. By leveraging \ac{DLT} for distributed open coordination, we aim to address the global cost-efficiency problem in market-based \ac{MRS} coordination~\cite{quintonMarketApproachesMultiRobot2023a}, advocating for creating a trustworthy cross-organizational platform. Section~\ref{sec:related_work} introduces the foundations of \ac{DLT}, including smart contracts and tokens, and reviews existing works on \ac{DLT} in \ac{MRS} and space. Section~\ref{sec:research_gap} details the technical challenges of using \ac{DLT} in space \ac{MRS}. Finally, Section~\ref{sec:4} presents a use case for distributed auction-based mapping exploration, focusing on the ESA-ESRIC Space Resource Challenge~\cite{linkESAESRICSpaceResources2021}. We conclude in Section~\ref{sec:5}.

\section{Related Work}
\label{sec:related_work}

We draw upon systematic literature reviews on market-based approaches in \ac{MRS}~\cite{diasMarketBasedMultirobotCoordination2006c, quintonMarketApproachesMultiRobot2023a}, research on space \ac{MRS}~\cite{gaoReviewSpaceRobotics2017}, and the application of \ac{DLT} in robotics and the space industry~\cite{aditya2021survey, ibrahimLiteratureReviewBlockchain2021}.

\ac{DLT} gained prominence with its first application, the cryptocurrency Bitcoin~\cite{Nakamoto:2008:Bitcoin}. As a subset of \ac{DLT}, blockchain technology is characterized by the replicated synchronized transaction processing and a decentralized consensus algorithm with different levels of participation, permissionless vs. permissioned, and access, public vs. private~\cite{butijnBlockchainsSystematicMultivocal2020b}. \ac{DLT} offers novel opportunities for digital interaction in a non-proprietary digital infrastructure, such as creating trust within consortiums and enabling new trading streams via Fungible and Non-Fungible Tokens (NFTs). NFTs can represent, among many other possibilities, ownership of an non-interchangeable asset, such as collectibles or art pieces~\cite{hartwichMachineEconomies2023}.
Besides the simple transfer of ownership, \ac{DLT} also allows implementing programming logic through uploading code known as smart contracts and invoking this code's exposed methods through transactions. In essence, smart contracts allow \ac{DLT} to provide the same functionality as a centralized platform, with two exceptions: first, non-deterministic methods are not supported, and second, the smart contract code, as well as all its inputs, intermediary results, and outputs, when triggered through a transaction are available to all \ac{DLT} nodes owing to the replicated execution and storage of transactions~\cite{kannengiesserChallengesCommonSolutions2022}.

Market-based approaches have been suggested for various applications in \ac{MRS}~\cite{quintonMarketApproachesMultiRobot2023a} and, specifically, space mapping~\cite{diasMarketBasedMultirobotCoordination2006c}. These approaches often leverage auction systems as suitable tools to efficiently decide which resources to use in a competitive space environment~\cite{zlotMultirobotExplorationControlled2002a}. Previous research suggests that robots can offer dependable services to each other by incorporating \ac{DLT}~\cite{lonshakovRobonomicsPlatformIntegration2018}, thereby opening up possibilities for space exploration, such as selling maps, facilitating decentralized decision-making, and enabling autonomous behavior. Integrating \ac{DLT} is also said to leverage agent cooperation and communication aimed at detection, learning, and autonomous penalization \cite{pittarasCooperativeReinforcementLearning2022}, improving heterogeneous robots' capabilities allowing them to perform complex tasks collaboratively.

Applications of \ac{DLT} in space have also been suggested for recording orbital positions, mining licenses, and managing space traffic~\cite{spaceConsenSysSpace2018}.
Another suggested application of \ac{DLT} in space is to support the efficient and effective coordination of \ac{MRS} missions, such as in-orbit operations~\cite{cordesDesignExperimentalEvaluation2018}, satellite formations, surface, and planetary exploration. By enabling robots to communicate and make transactions and decisions autonomously, \ac{DLT} is said to support the independent operation of individual robots and enable them to adapt to dynamic environments. 
Examples of single-entity deployments of \ac{DLT} that consider economic transactions include the AIRA~\cite{lonshakovRobonomicsPlatformIntegration2018} project, which introduces the concept of ``Robonomics,'' where heterogeneous robots and humans can contract and offer services and receive payment through smart contracts.

Despite advancements, earlier suggestions have focused on centralized single-entity settings. As such, they are not considering the main reason for using \ac{DLT} in a multi-participatory and competitive space environment and the novel challenges that may arise from it. In other words, all the platform functionalities that related work relies on could also be provided as a service through a dedicated provider that creates the needed digital infrastructure for one or multiple platforms in space~\cite{orlovaPresentFutureSpace2020} or uses a decentralized peer-to-peer communication network to create such a platform. Moreover, existing works still need to adequately address the need for flexible information representation and autonomous robot task learning in dynamic environments~\cite{baimaModelingObjectAffordances2021b}. A better understanding of the economic opportunities and technical challenges is indispensable to assess and implement \ac{DLT} in this context.

\section{Research Gap and Open Challenges}
\label{sec:research_gap}

Our discussion is grounded in existing research that tackles the challenges of \ac{DLT} from various angles, such as resource consumption, performance, and transparency~\cite{sedlmeir2022transparency, sedlmeir2020energy}. While \citet{aditya2021survey} have explored the general challenges that \ac{DLT} faces in robotics, we focus on five specific open problems particularly relevant to \ac{DLT} in space~\cite{ibrahimLiteratureReviewBlockchain2021} and market-based \ac{MRS}~\cite{diasMarketBasedMultirobotCoordination2006c, quintonMarketApproachesMultiRobot2023a}.

The first challenge is achieving high levels of autonomy in resource-limited environments, a recurring issue in space exploration~\cite{triantafyllouMethodologyApproachingIntegration2021a, lorenzHowFarFar2020}. While ontologies and semantic web have been proposed as a solution~\cite{memduhogluPossibleContributionsSpatial2018}, they often face difficulties dealing with data heterogeneity and system stability in resource-limited settings. Furthermore, establishing a fault-proof infrastructure and a trustless environment is crucial for autonomous coordination~\cite{quintonMarketApproachesMultiRobot2023a, orlovaPresentFutureSpace2020}. Given the constraints on storage and computational capacity in space robots due to the need for temperature and radiation hardening~\cite{gaoReviewSpaceRobotics2017}, thorough modeling and testing of \ac{DLT} in \ac{MRS} are essential. This includes evaluating public vs. private networks, permissionless vs. permissioned topologies, and possible \ac{ISRU}-specific consensus mechanisms, such as consensus-based auction and bundle algorithms~\cite{quintonMarketApproachesMultiRobot2023a}. The most advanced existing solutions for heterogenous \ac{MRS}, such as \cite{schusterARCHESSpaceAnalogueDemonstration2020} and \cite{dettmannCOROBXCooperativeRobot2022}, assume a fully cooperative system. Evaluating a coopetitive system remains an open problem.

The second challenge involves the universal cost/reward function for \ac{MRS} in market-based coordination~\cite{quintonMarketApproachesMultiRobot2023a}. Despite various proposals to improve bid valuation, such as distance-based metrics or performance indexes, the computational complexity of combinatorial auctions still needs to improve~\cite{diasMarketBasedMultirobotCoordination2006c}. Consequently, researchers have explored alternatives, such as clustering from single-item auctions to consensus-based bundle algorithms. However, current solutions often rely on proprietary frameworks, needing more transparency and trustless requirements for a digital platform in the emerging space economy~\cite{orlovaPresentFutureSpace2020}. To optimize \ac{ISRU} global efficiency, companies must consider outsourcing mapping data to other \ac{MRS} companies willing to sell instead of individually mapping it.

Thirdly, public narratives about blockchain technology are vital for its acceptance in governing outer space~\cite{filippiBlockchainOuterSpace2021}, and misconceptions about high energy consumption have been a barrier~\cite{sedlmeir2020energy}. The replicated processing of transactions storage using blockchain still increases the total resource consumption in terms of bandwidth, computing power, and electricity by a factor of $N$ in a blockchain network of $N$ nodes compared to a centralized platform, without taking necessary backups into account~\cite{sedlmeir2020energy}. Nevertheless, design decisions such as not using proof of work as a consensus mechanism and approaches like sharding, roll-ups, and off-chain payment channels can somewhat mitigate these inefficiencies~\cite{strobelManagingByzantineRobots2018, sedlmeir2020energy, lonshakovRobonomicsPlatformIntegration2018,  hartwichMachineEconomies2023}. Moreover, further investigation is needed to understand how to maintain trust and collaboration in a \emph{coopetitive} economy~\cite{ghobadiCoopetitiveKnowledgeSharing2011b}.

Fourth, for a comprehensive evaluation of opportunities and challenges, \ac{DLT} research should also investigate alternatives, such as solutions that propose a cloud-based system that could be used in spacecraft networks to address data complexity and heterogeneity~\cite{baldorApplyingCloudComputing2013}. These alternatives require further investigation, especially concerning latency sensitivity in \ac{DLT} solutions for satellite communication~\cite{ibrahimLiteratureReviewBlockchain2021}. Primarily, latency characteristics should be investigated, as \ac{DLT} is known to be highly latency sensitive despite the safety guarantees even in non-synchronous environments~\cite{guggenberger2021fabric, gervaisSecurityPerformanceProof2016b}.

Lastly, recent works have demonstrated the potential of using artificial intelligence (AI) agents to autonomously learn incremental affordances~\cite{baimaModelingObjectAffordances2021b} and achieve Pareto efficiency and improve equality and productivity in competitive and dynamic market simulations~\cite{zhengAIEconomistImproving2020}. While these results promise to promote universal shared efficient knowledge, there needs to be more comprehensive coverage and insights on autonomous economic interactions in the existing literature on machine economies in general~\cite{hartwichMachineEconomies2023} and \ac{DLT} in heterogeneous space \ac{MRS} in specific. \ac{DLT} has been suggested to manage heterogeneous space systems and efficiently support decentralized behavior~\cite{ferrer2018robochain}. The increasing interest from the scientific community and industry participants~\cite{weinzierlCommercialSpaceAge2021} also supports the need for further research on cost-effective \ac{MRS}. Traditional optimization techniques have also been used to achieve decentralized, shared economies without \ac{DLT}~\cite{pilgerstorferSelfAdaptiveLearningDecentralized2017}; further performance comparisons are needed to understand the approximation heuristics of decentralized combinatorial optimization systems.

\section{Research Plan}
\label{sec:4}

By outlining the methodology, objectives, and critical components of our research plan, grounded in a hypothetical scenario developed as part of the ESA-ESRIC space resource challenge~\cite{linkESAESRICSpaceResources2021}, we aim to explore the integration of blockchain technology extending to the proposed market-based \ac{MRS} for space exploration~\cite{diasMarketBasedMultirobotCoordination2006c}.

\begin{figure}
	\centering
	\includegraphics[width=0.47\textwidth]{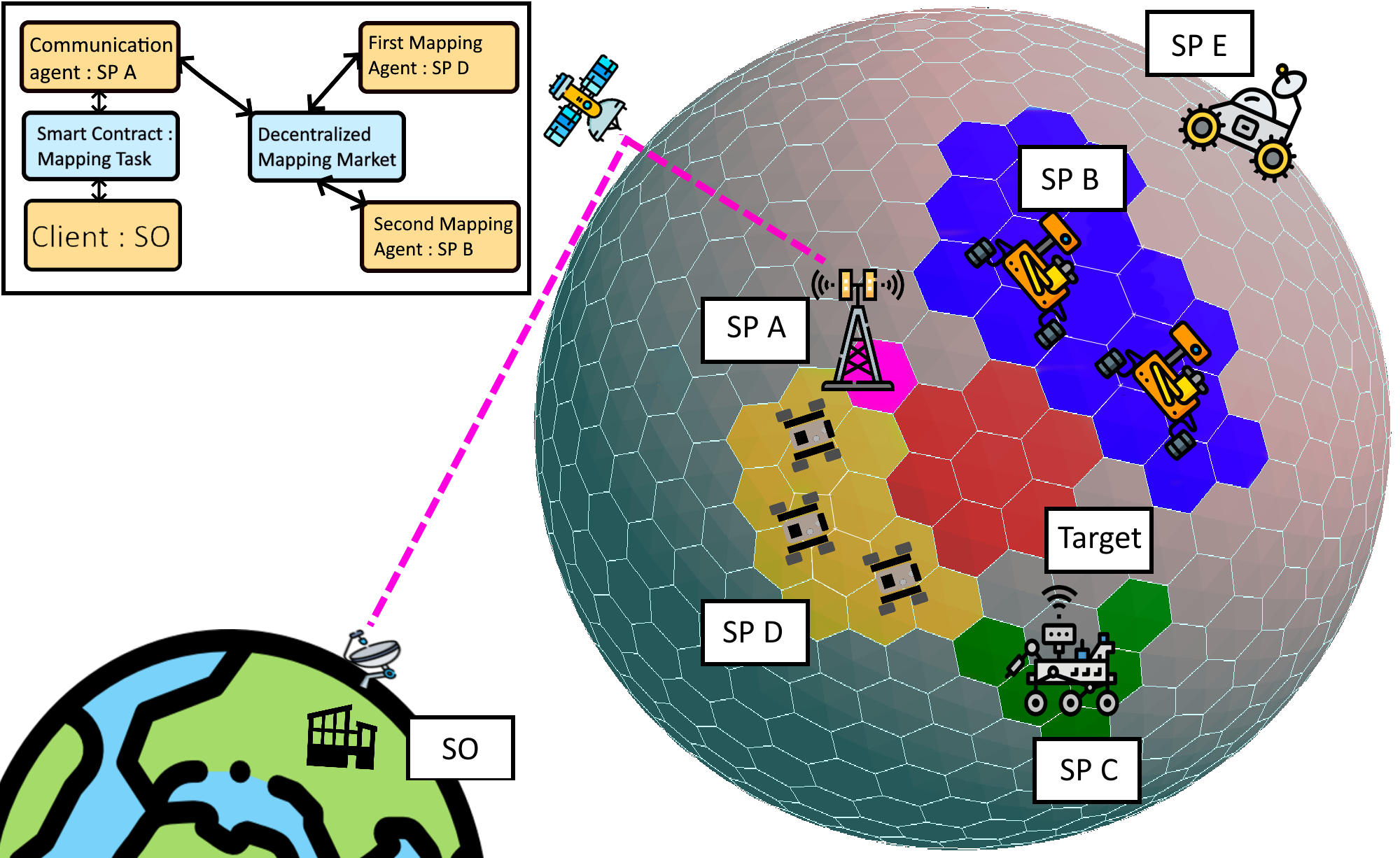}
	\caption{Scelestial mapping scenario, where each zone color represents a company's collaboration}
    \label{fig:Moonmap}
\end{figure}

In the given scenario, depicted in Figure~\ref{fig:Moonmap}, six main actors are involved: the Service Orderer (SO), an Earth-based organization responsible for mission directives, and five Service Providers (SP) -- entities that execute the tasks in-situ -- as roles elaborated in Table~\ref{tab:ListSP}. While the decentralized \ac{MRS} coordination concept dates back to the 1980s~\cite{millerSpatialRepresentationSystem1985, aubertCentralizedVsDecentralized1986}, recent surveys lacking related approaches~\cite{quintonMarketApproachesMultiRobot2023a, aditya2021survey} indicate that the full potential of \ac{DLT} in this domain still needs to be explored. To fill the research gaps, we prioritize in our plan the following essential requirements:

\begin{enumerate}[label=\roman*.]
\item \textbf{Network}: Enable a mesh network that lets clients create job postings and robots to communicate and coordinate trustfully via market-based tasks~\cite{diasMarketBasedMultirobotCoordination2006c, quintonMarketApproachesMultiRobot2023a}.
\item \textbf{Data Sharing Transparency}: Ensure transparency in data sharing to foster a \emph{coopetitive} environment with reduced information asymmetry~\cite{ghobadiCoopetitiveKnowledgeSharing2011b, lorenzHowFarFar2020}.
\item \textbf{Robot Agnostic System}: Develop a system compatible with any robotic platform and scalable to growing demands~\cite{diasMarketBasedMultirobotCoordination2006c}.
\item \textbf{Data Loss Resistance}: Implement safeguards to ensure data integrity and accessibility, even in the face of local system failures or network disruptions~\cite{orlovaPresentFutureSpace2020, nasaSpaceTechnologyRoadmaps2016, lorenzHowFarFar2020}.
\end{enumerate}

\begin{table}
\centering
\caption{List of service providers.}
\label{tab:ListSP}
\resizebox{\columnwidth}{!}{
\begin{tabular}{llllp{0.3\columnwidth}} 
\toprule
SP & Color & Robotic Fleet& \multicolumn{1}{c}{Main Focus} \\ 
\midrule
A & Pink & \begin{tabular}[t]{@{}l@{}}Communications satellites,\\ Multiple antennas\end{tabular} & \multicolumn{1}{l}{\begin{tabular}[t]{@{}l@{}}Moon-earth\\communications\end{tabular}} \\
B & Blue & Medium size offline fleet of robots & Mapping \\
C & Green & Few robots w/ embedded sensors & Resource analysis \\
D & Yellow & Large fleet of small robots & \multicolumn{1}{l}{\begin{tabular}[t]{@{}l@{}}Fast mapping with\\less precision\end{tabular}} \\
E & None & A single robot & Mapping \\
\bottomrule
\end{tabular}
}
\end{table}

In the scenario depicted in Figure~\ref{fig:Moonmap}, each polygon, structured using the Goldberg polyhedron, represents a sub-map owned and traded by network members~\cite{gooDIANABlockchainLunar2019}. This approach reduces the coordination complexity and allows for a fixed data size for each map segment, optimizing the data storage on the blockchain. However, because different sections of the map may represent distinct territories to be explored, ownership of the gathered information is transferred to the client rather than the explorer. When a robot creates data, it is desirable to represent it as an NFT to prove the authenticity of the knowledge produced and shared on the network. Such would, for instance, prevent unauthorized marketing as the original explorer, yet excessive information about the map would render its value because other \ac{DLT} nodes could retrieve it for free.

Our initial implementation involves a decentralized system of three rovers designed for resource identification, context analysis, and environmental mapping. These rovers have components comparable to space-graded hardware and can run blockchain nodes. The system uses a mesh network for communication and centralized data processing. However, as the \ac{MRS} for exploration design decision, the ultimate aim is to transition to a fully decentralized system supported by our successful simulations under similar Lunar network conditions. While our research plan offers a comprehensive approach to integrating \ac{DLT} with \ac{MRS}, as we argued in Section~\ref{sec:related_work}, a critical aspect of future research will be to focus on optimizing the throughput and resource consumption of blockchain nodes. The initial data organization strategy significantly enhanced data transmission efficiency, particularly regarding critical bandwidth usage. Yet, it is essential to investigate the specific conditions under which \ac{DLT} is most effective compared to authenticated and accountable bilateral communication based on digital signatures.

\section{Conclusion}
\label{sec:5}

This paper has presented a comprehensive research plan for integrating \Ac{DLT} with market-based heterogeneous \Ac{MRS} in space exploration. We have critically examined the existing literature to identify significant research gaps and articulated the necessity for interdisciplinary endeavors encompassing legal, regulatory, engineering, organizational, and economic aspects. Our proposed research agenda, while ambitious, aims to develop an \ac{MRS} architecture incorporating \ac{DLT} and enabling autonomous economic decision-making among robotic agents. We have also highlighted the unique technical challenges that arise when deploying \ac{DLT}-based systems in the harsh conditions of outer space. While the primary focus is on space applications, the principles, and architectures discussed have broader implications, potentially revolutionizing economic routes that could be adapted for terrestrial applications.

It is important to note that while this paper emphasizes practical application and feasibility, the challenges and use cases presented are not exhaustive. Our objective is to stimulate further research and discussion in this area, and we invite other scholars to extend, critique, or build upon our proposed framework. By laying the groundwork for future research, we hope to contribute to developing more efficient, transparent, and autonomous systems for space exploration. Designing such a system integrating \ac{DLT} with \ac{MRS} can significantly advance the field of software engineering, especially for robotics and automation software, offering new avenues for innovation and collaboration.

\begin{acks}
This research was funded in part by the Luxembourg National Research Fund (FNR) in the FiReSpARX (ref. 14783405) and PABLO (ref. 16326754) projects, and by PayPal, PEARL grant reference 13342933. For the purpose of open access, and in fulfillment of the obligations arising from the grant agreement, the author has applied a Creative Commons Attribution 4.0 International (CC BY 4.0) license to any Author Accepted Manuscript version arising from this submission.
\end{acks}

\bibliographystyle{ACM-Reference-Format}
\bibliography{references}


\begin{thebibliography}{50}

%%% the Web address.

\ifx \showCODEN    \undefined \def \showCODEN     #1{\unskip}     \fi
\ifx \showDOI      \undefined \def \showDOI       #1{#1}\fi
\ifx \showISBNx    \undefined \def \showISBNx     #1{\unskip}     \fi
\ifx \showISBNxiii \undefined \def \showISBNxiii  #1{\unskip}     \fi
\ifx \showISSN     \undefined \def \showISSN      #1{\unskip}     \fi
\ifx \showLCCN     \undefined \def \showLCCN      #1{\unskip}     \fi
\ifx \shownote     \undefined \def \shownote      #1{#1}          \fi
\ifx \showarticletitle \undefined \def \showarticletitle #1{#1}   \fi
\ifx \showURL      \undefined \def \showURL       {\relax}        \fi
% invisible to TeX
\providecommand\bibfield[2]{#2}
\providecommand\bibinfo[2]{#2}
\providecommand\natexlab[1]{#1}
\providecommand\showeprint[2][]{arXiv:#2}

\bibitem[Aditya et~al\mbox{.}(2021)]%
        {aditya2021survey}
\bibfield{author}{\bibinfo{person}{U.~S. P.~Srinivas Aditya},
  \bibinfo{person}{Roshan Singh}, \bibinfo{person}{Pranav~Kumar Singh}, {and}
  \bibinfo{person}{Anshuman Kalla}.} \bibinfo{year}{2021}\natexlab{}.
\newblock \showarticletitle{A Survey on Blockchain in Robotics: {{Issues}},
  Opportunities, Challenges and Future Directions}.
\newblock \bibinfo{journal}{\emph{Journal of Network and Computer
  Applications}}  \bibinfo{volume}{196} (\bibinfo{date}{Dec.}
  \bibinfo{year}{2021}), \bibinfo{pages}{10}.
\newblock
\showISSN{1084-8045}
\urldef\tempurl%
\url{https://doi.org/10.1016/j.jnca.2021.103245}
\showDOI{\tempurl}


\bibitem[Akerlof(1970)]%
        {akerlofMarketLemonsQuality1970}
\bibfield{author}{\bibinfo{person}{George~A. Akerlof}.}
  \bibinfo{year}{1970}\natexlab{}.
\newblock \showarticletitle{The Market for "{{Lemons}}": {{Quality}}
  Uncertainty and the Market Mechanism}.
\newblock \bibinfo{journal}{\emph{The Quarterly Journal of Economics}}
  \bibinfo{volume}{84}, \bibinfo{number}{3} (\bibinfo{date}{Aug.}
  \bibinfo{year}{1970}), \bibinfo{pages}{488--500}.
\newblock
\showISSN{0033-5533}
\urldef\tempurl%
\url{https://doi.org/10.2307/1879431}
\showDOI{\tempurl}


\bibitem[{Al-Akhar}(2019)]%
        {al-akharSpaceScienceTechnology2019}
\bibfield{author}{\bibinfo{person}{Rabi' {Al-Akhar}}.}
  \bibinfo{year}{2019}\natexlab{}.
\newblock \bibinfo{title}{Space Science and Technology - the Official Portal of
  the {{UAE}} Government}.
\newblock , \bibinfo{numpages}{14}~pages.
\newblock
\urldef\tempurl%
\url{https://u.ae/en/about-the-uae/science-and-technology/key-sectors-in-science-and-technology/space-science-and-techno}
\showURL{%
\tempurl}


\bibitem[Aubert et~al\mbox{.}(1986)]%
        {aubertCentralizedVsDecentralized1986}
\bibfield{author}{\bibinfo{person}{Ph.~Saint Aubert}, \bibinfo{person}{M.
  Hervieux}, \bibinfo{person}{J.~L. Perbos}, \bibinfo{person}{E. Saggese},
  {and} \bibinfo{person}{C. Soprano}.} \bibinfo{year}{1986}\natexlab{}.
\newblock \showarticletitle{Centralized vs Decentralized Options for a European
  Data Relay Satellite System}.
\newblock \bibinfo{journal}{\emph{Acta Astronautica}}  \bibinfo{volume}{13}
  (\bibinfo{year}{1986}), \bibinfo{pages}{387}.
\newblock
\showISSN{0094-5765}
\urldef\tempurl%
\url{https://doi.org/10.1016/0094-5765(86)90093-7}
\showDOI{\tempurl}


\bibitem[Baima and Colombini(2021)]%
        {baimaModelingObjectAffordances2021b}
\bibfield{author}{\bibinfo{person}{Renan~Lima Baima} {and}
  \bibinfo{person}{Esther~Luna Colombini}.} \bibinfo{year}{2021}\natexlab{}.
\newblock \showarticletitle{Modeling Object's Affordances via Reward
  Functions}. In \bibinfo{booktitle}{\emph{Proc. {{Int}}. {{Conf}}. on
  {{Sys}}., {{Man}}, and {{Cyb}}.}} \bibinfo{publisher}{{IEEE}},
  \bibinfo{address}{{Melbourne, Australia}}, \bibinfo{pages}{2183--2190}.
\newblock
\showISBNx{978-1-66544-207-7}
\urldef\tempurl%
\url{https://doi.org/10.1109/smc52423.2021.9658915}
\showDOI{\tempurl}


\bibitem[Baldor et~al\mbox{.}(2013)]%
        {baldorApplyingCloudComputing2013}
\bibfield{author}{\bibinfo{person}{Sue~A. Baldor}, \bibinfo{person}{Carlos
  Quiroz}, {and} \bibinfo{person}{Paul Wood}.} \bibinfo{year}{2013}\natexlab{}.
\newblock \showarticletitle{Applying a Cloud Computing Approach to Storage
  Architectures for Spacecraft}. In \bibinfo{booktitle}{\emph{2013 {{IEEE
  Aerospace Conference}}}}. \bibinfo{publisher}{{IEEE}}, \bibinfo{address}{{Big
  Sky, Montana}}, \bibinfo{pages}{1--6}.
\newblock
\showISBNx{978-1-4673-1813-6}
\urldef\tempurl%
\url{https://doi.org/10.1109/aero.2013.6497340}
\showDOI{\tempurl}


\bibitem[Borowitz(2019)]%
        {borowitzStrategicImplicationsProliferation2019}
\bibfield{author}{\bibinfo{person}{Mariel Borowitz}.}
  \bibinfo{year}{2019}\natexlab{}.
\newblock \showarticletitle{Strategic Implications of the Proliferation of
  Space Situational Awareness Technology and Information: Lessons Learned from
  the Remote Sensing Sector}.
\newblock \bibinfo{journal}{\emph{Space Policy}}  \bibinfo{volume}{47}
  (\bibinfo{date}{Feb.} \bibinfo{year}{2019}), \bibinfo{pages}{18--27}.
\newblock
\showISSN{0265-9646}
\urldef\tempurl%
\url{https://doi.org/10.1016/j.spacepol.2018.05.002}
\showDOI{\tempurl}


\bibitem[Butijn et~al\mbox{.}(2020)]%
        {butijnBlockchainsSystematicMultivocal2020b}
\bibfield{author}{\bibinfo{person}{Bert-Jan Butijn}, \bibinfo{person}{Damian~A.
  Tamburri}, {and} \bibinfo{person}{Willem-Jan van~den Heuvel}.}
  \bibinfo{year}{2020}\natexlab{}.
\newblock \showarticletitle{Blockchains: {{A}} Systematic Multivocal Literature
  Review}.
\newblock \bibinfo{journal}{\emph{Comput. Surveys}} \bibinfo{volume}{53},
  \bibinfo{number}{3}, Article \bibinfo{articleno}{61} (\bibinfo{date}{July}
  \bibinfo{year}{2020}), \bibinfo{numpages}{37}~pages.
\newblock
\showISSN{0360-0300}
\urldef\tempurl%
\url{https://doi.org/10.1145/3369052}
\showDOI{\tempurl}


\bibitem[Conn(2021)]%
        {connGlobalSpaceEconomy2021}
\bibfield{author}{\bibinfo{person}{Lesley Conn}.}
  \bibinfo{year}{2021}\natexlab{}.
\newblock \bibinfo{title}{Global Space Economy Nears \${{447B}}}.
\newblock
\newblock
\urldef\tempurl%
\url{https://www.thespacereport.org/uncategorized/global-space-economy-nears-447b/}
\showURL{%
\tempurl}


\bibitem[{Consensys Space}(2018)]%
        {spaceConsenSysSpace2018}
\bibfield{author}{\bibinfo{person}{{Consensys Space}}.}
  \bibinfo{year}{2018}\natexlab{}.
\newblock \bibinfo{title}{Open {{Source Space}}}.
\newblock
\newblock
\urldef\tempurl%
\url{https://www.consensys.space/}
\showURL{%
\tempurl}


\bibitem[Cordes(2018)]%
        {cordesDesignExperimentalEvaluation2018}
\bibfield{author}{\bibinfo{person}{Florian Cordes}.}
  \bibinfo{year}{2018}\natexlab{}.
\newblock \emph{\bibinfo{title}{Design and Experimental Evaluation of a Hybrid
  Wheeled-Leg Exploration Rover in the Context of Multi-Robot Systems}}.
\newblock Kumulative Dissertation. \bibinfo{school}{Universit{\"a}t Bremen},
  \bibinfo{address}{{Bremen, Germany}}.
\newblock


\bibitem[Crawford(2015)]%
        {crawfordLunarResourcesReview2015}
\bibfield{author}{\bibinfo{person}{Ian~A. Crawford}.}
  \bibinfo{year}{2015}\natexlab{}.
\newblock \showarticletitle{Lunar Resources: {{A}} Review}.
\newblock \bibinfo{journal}{\emph{Progress in Physical Geography: Earth}}
  \bibinfo{volume}{39}, \bibinfo{number}{2} (\bibinfo{date}{April}
  \bibinfo{year}{2015}), \bibinfo{pages}{137--167}.
\newblock
\showISSN{0309-1333}
\urldef\tempurl%
\url{https://doi.org/10.1177/0309133314567585}
\showDOI{\tempurl}


\bibitem[De~Filippi and Leiter(2021)]%
        {filippiBlockchainOuterSpace2021}
\bibfield{author}{\bibinfo{person}{P. De~Filippi} {and} \bibinfo{person}{Andrea
  Leiter}.} \bibinfo{year}{2021}\natexlab{}.
\newblock \showarticletitle{Blockchain in Outer Space}.
\newblock \bibinfo{journal}{\emph{American Journal of International Law}}
  \bibinfo{volume}{115} (\bibinfo{year}{2021}), \bibinfo{pages}{413--418}.
\newblock
\showISSN{2398-7723}
\urldef\tempurl%
\url{https://doi.org/10.1017/aju.2021.63}
\showDOI{\tempurl}


\bibitem[Deshpande and Luntz(2003)]%
        {deshpandeDecentralizedControlTeam2003}
\bibfield{author}{\bibinfo{person}{A. Deshpande} {and} \bibinfo{person}{J.
  Luntz}.} \bibinfo{year}{2003}\natexlab{}.
\newblock \showarticletitle{Decentralized Control for a Team of Physically
  Cooperating Robots}. In \bibinfo{booktitle}{\emph{Proc. {{Int}}. {{Conf}}.
  {{Intell}}. {{Robots Syst}}.}}, Vol.~\bibinfo{volume}{2}.
  \bibinfo{publisher}{{IEEE}}, \bibinfo{address}{{Las Vegas, Nevada, USA}},
  \bibinfo{pages}{1757--1762 vol.2}.
\newblock
\urldef\tempurl%
\url{https://doi.org/10.1109/iros.2003.1248898}
\showDOI{\tempurl}


\bibitem[Dettmann et~al\mbox{.}(2022)]%
        {dettmannCOROBXCooperativeRobot2022}
\bibfield{author}{\bibinfo{person}{Alexander Dettmann}, \bibinfo{person}{Thomas
  Voegele}, \bibinfo{person}{Jorge Oc{\'o}n}, \bibinfo{person}{Iulia Dragomir},
  \bibinfo{person}{Shashank Govindaraj}, \bibinfo{person}{Matteo {de
  Benedetti}}, \bibinfo{person}{Val{\'e}rie Ciarletti}, \bibinfo{person}{Rafik
  {Hassen-Khodja}}, \bibinfo{person}{Thierry Germa}, \bibinfo{person}{Raphael
  Viards}, \bibinfo{person}{Gonzalo~J. {Paz-Delgado}}, {and}
  \bibinfo{person}{Laura~M. Mantoani}.} \bibinfo{year}{2022}\natexlab{}.
\newblock \showarticletitle{{{COROB-X}}: A Cooperative Robot Team for the
  Exploration of Lunar Skylights}. In \bibinfo{booktitle}{\emph{{{ASTRA}} 2022
  16th {{Symposium}} on {{Advanced Space Technologies}} in {{Robotics}} and
  {{Automation}}}}. \bibinfo{publisher}{{ESA-ESTEC}},
  \bibinfo{address}{{Noordwijk, Netherlands}}, \bibinfo{pages}{9}.
\newblock
\urldef\tempurl%
\url{https://hal-insu.archives-ouvertes.fr/insu-03751549}
\showURL{%
\tempurl}


\bibitem[Dias et~al\mbox{.}(2006)]%
        {diasMarketBasedMultirobotCoordination2006c}
\bibfield{author}{\bibinfo{person}{M.~Bernardine Dias}, \bibinfo{person}{Robert
  Zlot}, \bibinfo{person}{Nidhi Kalra}, {and} \bibinfo{person}{Anthony
  Stentz}.} \bibinfo{year}{2006}\natexlab{}.
\newblock \showarticletitle{Market-{{Based Multirobot Coordination}}: {{A
  Survey}} and {{Analysis}}}.
\newblock \bibinfo{journal}{\emph{Proc. IEEE}} \bibinfo{volume}{94},
  \bibinfo{number}{7} (\bibinfo{date}{July} \bibinfo{year}{2006}),
  \bibinfo{pages}{1257--1270}.
\newblock
\showISSN{0018-9219, 1558-2256}
\urldef\tempurl%
\url{https://doi.org/10.1109/JPROC.2006.876939}
\showDOI{\tempurl}


\bibitem[Ferrer et~al\mbox{.}(2018)]%
        {ferrer2018robochain}
\bibfield{author}{\bibinfo{person}{Eduardo~Castell{\'o} Ferrer},
  \bibinfo{person}{Ognjen Rudovic}, \bibinfo{person}{Thomas Hardjono}, {and}
  \bibinfo{person}{Alex Pentland}.} \bibinfo{year}{2018}\natexlab{}.
\newblock \showarticletitle{{{RoboChain}}: A Secure Data-Sharing Framework for
  Human-Robot Interaction}. In \bibinfo{booktitle}{\emph{The {{Tenth
  International Conference}} on {{eHealth}}, {{Telemedicine}}, and {{Social
  Medicine}}}}. \bibinfo{publisher}{{IARA XPS Press}}, \bibinfo{address}{{Rome,
  Italy}}, \bibinfo{pages}{124 to 130}.
\newblock
\showISBNx{978-1-61208-618-7}


\bibitem[Gao and Chien(2017)]%
        {gaoReviewSpaceRobotics2017}
\bibfield{author}{\bibinfo{person}{Yang Gao} {and} \bibinfo{person}{Steve
  Chien}.} \bibinfo{year}{2017}\natexlab{}.
\newblock \showarticletitle{Review on Space Robotics: {{Toward}} Top-Level
  Science through Space Exploration}.
\newblock \bibinfo{journal}{\emph{Science Robotics}} \bibinfo{volume}{2},
  \bibinfo{number}{7} (\bibinfo{date}{June} \bibinfo{year}{2017}),
  \bibinfo{pages}{eaan5074}.
\newblock
\urldef\tempurl%
\url{https://doi.org/10.1126/scirobotics.aan5074}
\showDOI{\tempurl}


\bibitem[Garc{\i}a et~al\mbox{.}(2018)]%
        {garciaArchitectureDecentralizedCollaborative2018}
\bibfield{author}{\bibinfo{person}{Sergio Garc{\i}a}, \bibinfo{person}{Claudio
  Menghi}, \bibinfo{person}{Patrizio Pelliccione}, \bibinfo{person}{Thorsten
  Berger}, {and} \bibinfo{person}{Rebekka Wohlrab}.}
  \bibinfo{year}{2018}\natexlab{}.
\newblock \showarticletitle{An {{Architecture}} for {{Decentralized}},
  {{Collaborative}}, and {{Autonomous Robots}}}. In
  \bibinfo{booktitle}{\emph{2018 {{IEEE International Conference}} on
  {{Software Architecture}} ({{ICSA}})}}. \bibinfo{publisher}{{IEEE}},
  \bibinfo{address}{{Seattle, WA}}, \bibinfo{pages}{75--7509}.
\newblock
\showISBNx{978-1-5386-6398-1}
\urldef\tempurl%
\url{https://doi.org/10.1109/ICSA.2018.00017}
\showDOI{\tempurl}


\bibitem[Gervais et~al\mbox{.}(2016)]%
        {gervaisSecurityPerformanceProof2016b}
\bibfield{author}{\bibinfo{person}{Arthur Gervais}, \bibinfo{person}{Ghassan~O.
  Karame}, \bibinfo{person}{Karl W{\"u}st}, \bibinfo{person}{Vasileios
  Glykantzis}, \bibinfo{person}{Hubert Ritzdorf}, {and} \bibinfo{person}{Srdjan
  Capkun}.} \bibinfo{year}{2016}\natexlab{}.
\newblock \showarticletitle{On the Security and Performance of Proof of Work
  Blockchains}. In \bibinfo{booktitle}{\emph{Proceedings of the {{ACM SIGSAC
  Conference}} on {{Computer}} and {{Communications Security}}}}
  \emph{(\bibinfo{series}{{{CCS}} '16})}. \bibinfo{publisher}{{Association for
  Computing Machinery}}, \bibinfo{address}{{New York, NY, USA}},
  \bibinfo{pages}{3--16}.
\newblock
\showISBNx{978-1-4503-4139-4}
\urldef\tempurl%
\url{https://doi.org/10.1145/2976749.2978341}
\showDOI{\tempurl}


\bibitem[Ghobadi and D'Ambra(2011)]%
        {ghobadiCoopetitiveKnowledgeSharing2011b}
\bibfield{author}{\bibinfo{person}{Shahla Ghobadi} {and} \bibinfo{person}{John
  D'Ambra}.} \bibinfo{year}{2011}\natexlab{}.
\newblock \showarticletitle{Coopetitive Knowledge Sharing: {{An}} Analytical
  Review of Literature}.
\newblock \bibinfo{journal}{\emph{Electronic Journal of Knowledge Management}}
  \bibinfo{volume}{9}, \bibinfo{number}{4} (\bibinfo{date}{Jan.}
  \bibinfo{year}{2011}), \bibinfo{pages}{307--317}.
\newblock
\urldef\tempurl%
\url{https://research.manchester.ac.uk/en/publications/coopetitive-knowledge-sharing-an-analytical-review-of-literature}
\showURL{%
\tempurl}


\bibitem[Goo et~al\mbox{.}(2019)]%
        {gooDIANABlockchainLunar2019}
\bibfield{author}{\bibinfo{person}{Jason Goo}, \bibinfo{person}{Ojas Bora},
  \bibinfo{person}{Priyansh Manne}, \bibinfo{person}{Kazem Mullah},
  \bibinfo{person}{Kayden Tilden}, \bibinfo{person}{Jeremy Hartley},
  \bibinfo{person}{Ayame Kunieda}, \bibinfo{person}{Savannah Laver},
  \bibinfo{person}{Stephanie Lee}, \bibinfo{person}{Laquan Shuai}, {and}
  \bibinfo{person}{Isaiah Li~Yun~Tan}.} \bibinfo{year}{2019}\natexlab{}.
\newblock \bibinfo{title}{Blockchain Lunar Registry}.
\newblock
\newblock
\urldef\tempurl%
\url{https://diana.io}
\showURL{%
\tempurl}


\bibitem[Guggenberger et~al\mbox{.}(2022)]%
        {guggenberger2021fabric}
\bibfield{author}{\bibinfo{person}{Tobias Guggenberger},
  \bibinfo{person}{Johannes Sedlmeir}, \bibinfo{person}{Gilbert Fridgen}, {and}
  \bibinfo{person}{Andr{\'e} Luckow}.} \bibinfo{year}{2022}\natexlab{}.
\newblock \showarticletitle{An In-Depth Investigation of the Performance
  Characteristics of Hyperledger Fabric}.
\newblock \bibinfo{journal}{\emph{Computers \& Industrial Engineering}}
  \bibinfo{volume}{173} (\bibinfo{date}{Nov.} \bibinfo{year}{2022}),
  \bibinfo{pages}{108716}.
\newblock
\showISSN{0360-8352}
\urldef\tempurl%
\url{https://doi.org/10.1016/j.cie.2022.108716}
\showDOI{\tempurl}


\bibitem[Harrison and St.~John(1996)]%
        {harrisonManagingPartneringExternal1996a}
\bibfield{author}{\bibinfo{person}{Jeffrey~S. Harrison} {and}
  \bibinfo{person}{Caron~H. St.~John}.} \bibinfo{year}{1996}\natexlab{}.
\newblock \showarticletitle{Managing and Partnering with External
  Stakeholders}.
\newblock \bibinfo{journal}{\emph{Academy of Management Perspectives}}
  \bibinfo{volume}{10}, \bibinfo{number}{2} (\bibinfo{date}{May}
  \bibinfo{year}{1996}), \bibinfo{pages}{46--60}.
\newblock
\showISSN{1558-9080}
\urldef\tempurl%
\url{https://doi.org/10.5465/ame.1996.9606161554}
\showDOI{\tempurl}


\bibitem[Hartwich et~al\mbox{.}(2023)]%
        {hartwichMachineEconomies2023}
\bibfield{author}{\bibinfo{person}{Eduard Hartwich}, \bibinfo{person}{Alexander
  Rieger}, \bibinfo{person}{Johannes Sedlmeir}, \bibinfo{person}{Dominik
  Jurek}, {and} \bibinfo{person}{Gilbert Fridgen}.}
  \bibinfo{year}{2023}\natexlab{}.
\newblock \showarticletitle{Machine Economies}.
\newblock \bibinfo{journal}{\emph{Electronic Markets}} \bibinfo{volume}{33},
  \bibinfo{number}{1} (\bibinfo{date}{July} \bibinfo{year}{2023}),
  \bibinfo{pages}{36}.
\newblock
\showISSN{1422-8890}
\urldef\tempurl%
\url{https://doi.org/10.1007/s12525-023-00649-0}
\showDOI{\tempurl}


\bibitem[Ibrahim et~al\mbox{.}(2021)]%
        {ibrahimLiteratureReviewBlockchain2021}
\bibfield{author}{\bibinfo{person}{{\relax Hussein}. Ibrahim},
  \bibinfo{person}{Marwa~A. Shouman}, \bibinfo{person}{Nawal~A. {El-Fishawy}},
  {and} \bibinfo{person}{{\relax Ayman}. Ahmed}.}
  \bibinfo{year}{2021}\natexlab{}.
\newblock \showarticletitle{Literature Review of Blockchain Technology in Space
  Industry: Challenges and Applications}. In \bibinfo{booktitle}{\emph{Int.
  {{Conf}}. {{Electron}}. {{Eng}}.}} \bibinfo{publisher}{{IEEE}},
  \bibinfo{address}{{Menouf, Egypt}}, \bibinfo{pages}{1--8}.
\newblock
\showISBNx{978-1-66541-842-3}
\showLCCN{20876800}
\urldef\tempurl%
\url{https://doi.org/10.1109/ICEEM52022.2021.9480642}
\showDOI{\tempurl}


\bibitem[Kannengiesser et~al\mbox{.}(2022)]%
        {kannengiesserChallengesCommonSolutions2022}
\bibfield{author}{\bibinfo{person}{Niclas Kannengiesser},
  \bibinfo{person}{Sebastian Lins}, \bibinfo{person}{Christian Sander},
  \bibinfo{person}{Klaus Winter}, \bibinfo{person}{Hellmuth Frey}, {and}
  \bibinfo{person}{Ali Sunyaev}.} \bibinfo{year}{2022}\natexlab{}.
\newblock \showarticletitle{Challenges and Common Solutions in Smart Contract
  Development}.
\newblock \bibinfo{journal}{\emph{IEEE Transactions on Software Engineering}}
  \bibinfo{volume}{48}, \bibinfo{number}{11} (\bibinfo{date}{Dec.}
  \bibinfo{year}{2022}), \bibinfo{pages}{4291--4318}.
\newblock
\showISSN{1939-3520}
\urldef\tempurl%
\url{https://doi.org/10.1109/TSE.2021.3116808}
\showDOI{\tempurl}


\bibitem[Khan et~al\mbox{.}(2019)]%
        {khanBlockchainTechnologyApplications2019}
\bibfield{author}{\bibinfo{person}{Ameer~Tamoor Khan}, \bibinfo{person}{Xinwei
  Cao}, \bibinfo{person}{Shuai Li}, {and} \bibinfo{person}{Zoran Milosevic}.}
  \bibinfo{year}{2019}\natexlab{}.
\newblock \showarticletitle{Blockchain Technology with Applications to
  Distributed Control and Cooperative Robotics: {{A}} Survey}.
\newblock \bibinfo{journal}{\emph{International Journal of Robotics and
  Control}} \bibinfo{volume}{2}, \bibinfo{number}{1} (\bibinfo{date}{Jan.}
  \bibinfo{year}{2019}), \bibinfo{pages}{36}.
\newblock
\showISSN{2577-7769, 2577-7742}
\urldef\tempurl%
\url{https://doi.org/10.5430/ijrc.v2n1p36}
\showDOI{\tempurl}


\bibitem[Leon(2018)]%
        {leonMiningMeaningExamination2018}
\bibfield{author}{\bibinfo{person}{Amanda~M. Leon}.}
  \bibinfo{year}{2018}\natexlab{}.
\newblock \showarticletitle{Mining for Meaning: {{An}} Examination of the
  Legality of Property Rights in Space Resources}.
\newblock \bibinfo{journal}{\emph{Virginia Law Review}} \bibinfo{volume}{104},
  \bibinfo{number}{3} (\bibinfo{date}{May} \bibinfo{year}{2018}),
  \bibinfo{pages}{497--546}.
\newblock
\showISSN{00426601}
\showeprint[jstor]{44864188}
\urldef\tempurl%
\url{https://www.jstor.org/stable/44864188}
\showURL{%
\tempurl}


\bibitem[Link et~al\mbox{.}(2021)]%
        {linkESAESRICSpaceResources2021}
\bibfield{author}{\bibinfo{person}{Mathias Link}, \bibinfo{person}{David
  Parker}, \bibinfo{person}{Massimo Sabbatini}, {and}
  \bibinfo{person}{Franziska  }.} \bibinfo{year}{2021}\natexlab{}.
\newblock \bibinfo{title}{{{ESA-ESRIC}} Space Resources Challenge - Prospecting
  Technologies - Call}.
\newblock
\newblock
\showLCCN{1}
\urldef\tempurl%
\url{https://www.spaceresourceschallenge.esa.int}
\showURL{%
\tempurl}


\bibitem[Lonshakov et~al\mbox{.}(2018)]%
        {lonshakovRobonomicsPlatformIntegration2018}
\bibfield{author}{\bibinfo{person}{Sergey Lonshakov},
  \bibinfo{person}{Aleksandr Krupenkin}, \bibinfo{person}{Aleksandr Kapitonov},
  \bibinfo{person}{Evgeny Radchenko}, \bibinfo{person}{Alisher Khassanov},
  {and} \bibinfo{person}{A. Starostin}.} \bibinfo{year}{2018}\natexlab{}.
\newblock \bibinfo{title}{Robonomics: {{Platform}} for Integration of Cyber
  Physical Systems into Human Economy for Engineers, Smart Cities and
  {{Industry}} 4.0 Creators}.
\newblock
\newblock
\urldef\tempurl%
\url{https://doi.org/10.13140/RG.2.2.23928.60169}
\showDOI{\tempurl}


\bibitem[Lorenz(2020)]%
        {lorenzHowFarFar2020}
\bibfield{author}{\bibinfo{person}{Ralph~D. Lorenz}.}
  \bibinfo{year}{2020}\natexlab{}.
\newblock \showarticletitle{How Far Is Far Enough? {{Requirements}} Derivation
  for Planetary Mobility Systems}.
\newblock \bibinfo{journal}{\emph{Advances in Space Research}}
  \bibinfo{volume}{65}, \bibinfo{number}{5} (\bibinfo{date}{March}
  \bibinfo{year}{2020}), \bibinfo{pages}{1383--1401}.
\newblock
\showISSN{0273-1177}
\urldef\tempurl%
\url{https://doi.org/10.1016/j.asr.2019.12.011}
\showDOI{\tempurl}


\bibitem[Martinez~Rocamora et~al\mbox{.}(2023)]%
        {martinezrocamoraMultirobotCooperationLunar2023}
\bibfield{author}{\bibinfo{person}{Bernardo Martinez~Rocamora},
  \bibinfo{person}{Cagri Kilic}, \bibinfo{person}{Christopher Tatsch},
  \bibinfo{person}{Guilherme A.~S. Pereira}, {and} \bibinfo{person}{Jason~N.
  Gross}.} \bibinfo{year}{2023}\natexlab{}.
\newblock \showarticletitle{Multi-Robot Cooperation for Lunar {{In-Situ}}
  Resource Utilization}.
\newblock \bibinfo{journal}{\emph{Frontiers in Robotics and AI}}
  \bibinfo{volume}{10} (\bibinfo{date}{March} \bibinfo{year}{2023}),
  \bibinfo{pages}{1149080}.
\newblock
\showISSN{2296-9144}
\urldef\tempurl%
\url{https://doi.org/10.3389/frobt.2023.1149080}
\showDOI{\tempurl}


\bibitem[Memduhoglu and Basaraner(2018)]%
        {memduhogluPossibleContributionsSpatial2018}
\bibfield{author}{\bibinfo{person}{Abdulkadir Memduhoglu} {and}
  \bibinfo{person}{Melih Basaraner}.} \bibinfo{year}{2018}\natexlab{}.
\newblock \showarticletitle{Possible Contributions of Spatial Semantic Methods
  and Technologies to Multi-Representation Spatial Database Paradigm}.
\newblock \bibinfo{journal}{\emph{International Journal of Engineering and
  Geosciences}} \bibinfo{volume}{3}, \bibinfo{number}{3} (\bibinfo{date}{Oct.}
  \bibinfo{year}{2018}), \bibinfo{pages}{108--118}.
\newblock
\urldef\tempurl%
\url{https://doi.org/10.26833/ijeg.413473}
\showDOI{\tempurl}


\bibitem[Miller(1985)]%
        {millerSpatialRepresentationSystem1985}
\bibfield{author}{\bibinfo{person}{D. Miller}.}
  \bibinfo{year}{1985}\natexlab{}.
\newblock \showarticletitle{A Spatial Representation System for Mobile Robots}.
  In \bibinfo{booktitle}{\emph{1985 {{IEEE International Conference}} on
  {{Robotics}} and {{Automation Proceedings}}}}, Vol.~\bibinfo{volume}{2}.
  \bibinfo{publisher}{{IEEE}}, \bibinfo{address}{{St. Louis, MO, USA}},
  \bibinfo{pages}{122--127}.
\newblock
\urldef\tempurl%
\url{https://doi.org/10.1109/ROBOT.1985.1087318}
\showDOI{\tempurl}


\bibitem[Nakamoto(2008)]%
        {Nakamoto:2008:Bitcoin}
\bibfield{author}{\bibinfo{person}{Satoshi Nakamoto}.}
  \bibinfo{year}{2008}\natexlab{}.
\newblock \bibinfo{title}{Bitcoin: A Peer-to-Peer Electronic Cash System}.
\newblock
\newblock
\urldef\tempurl%
\url{https://bitcoin.org/bitcoin.pdf}
\showURL{%
\tempurl}


\bibitem[{NASA}(2016)]%
        {nasaSpaceTechnologyRoadmaps2016}
\bibfield{author}{\bibinfo{person}{{NASA}}.} \bibinfo{year}{2016}\natexlab{}.
\newblock \bibinfo{booktitle}{\emph{Space Technology Roadmaps and Priorities
  Revisited}}.
\newblock \bibinfo{type}{{T}echnical {R}eport}. \bibinfo{institution}{{The
  National Academies Press}}, \bibinfo{address}{{DC, USA}}.
\newblock
\urldef\tempurl%
\url{https://doi.org/10.17226/23582}
\showDOI{\tempurl}


\bibitem[Orlova et~al\mbox{.}(2020)]%
        {orlovaPresentFutureSpace2020}
\bibfield{author}{\bibinfo{person}{Alina Orlova}, \bibinfo{person}{Roberto
  Nogueira}, {and} \bibinfo{person}{Paula Chimenti}.}
  \bibinfo{year}{2020}\natexlab{}.
\newblock \showarticletitle{The {{Present}} and {{Future}} of the {{Space
  Sector}}: {{A Business Ecosystem Approach}}}.
\newblock \bibinfo{journal}{\emph{Space Policy}}  \bibinfo{volume}{52}
  (\bibinfo{date}{May} \bibinfo{year}{2020}), \bibinfo{pages}{101374}.
\newblock
\showISSN{0265-9646}
\urldef\tempurl%
\url{https://doi.org/10.1016/j.spacepol.2020.101374}
\showDOI{\tempurl}


\bibitem[Pilgerstorfer and Pournaras(2017)]%
        {pilgerstorferSelfAdaptiveLearningDecentralized2017}
\bibfield{author}{\bibinfo{person}{Peter Pilgerstorfer} {and}
  \bibinfo{person}{Evangelos Pournaras}.} \bibinfo{year}{2017}\natexlab{}.
\newblock \showarticletitle{Self-Adaptive Learning in Decentralized
  Combinatorial Optimization - {{A}} Design Paradigm for Sharing Economies}. In
  \bibinfo{booktitle}{\emph{2017 {{IEEE}}/{{ACM}} 12th {{International
  Symposium}} on {{Software Engineering}} for {{Adaptive}} and {{Self-Managing
  Systems}} ({{SEAMS}})}}. \bibinfo{publisher}{{IEEE}},
  \bibinfo{address}{{Buenos Aires, Argentina}}, \bibinfo{pages}{54--64}.
\newblock
\urldef\tempurl%
\url{https://doi.org/10.1109/seams.2017.8}
\showDOI{\tempurl}


\bibitem[Pittaras(2022)]%
        {pittarasCooperativeReinforcementLearning2022}
\bibfield{author}{\bibinfo{person}{Nikiforos Pittaras}.}
  \bibinfo{year}{2022}\natexlab{}.
\newblock \bibinfo{title}{A Cooperative Reinforcement Learning Environment for
  Detecting and Penalizing Betrayal}.
\newblock
\newblock
\urldef\tempurl%
\url{https://doi.org/10.48550/arXiv.2210.12841}
\showDOI{\tempurl}
\showeprint[arxiv]{2210.12841}


\bibitem[Pretto et~al\mbox{.}(2021)]%
        {prettoBuildingAerialGroundRobotics2021}
\bibfield{author}{\bibinfo{person}{Alberto Pretto},
  \bibinfo{person}{St{\'e}phanie Aravecchia}, \bibinfo{person}{Wolfram
  Burgard}, \bibinfo{person}{Nived Chebrolu}, \bibinfo{person}{Christian
  Dornhege}, \bibinfo{person}{Tillmann Falck}, \bibinfo{person}{Freya
  Fleckenstein}, \bibinfo{person}{Alessandra Fontenla}, \bibinfo{person}{Marco
  Imperoli}, \bibinfo{person}{Raghav Khanna}, \bibinfo{person}{Frank Liebisch},
  \bibinfo{person}{Philipp Lottes}, \bibinfo{person}{Andres Milioto},
  \bibinfo{person}{Daniele Nardi}, \bibinfo{person}{Sandro Nardi},
  \bibinfo{person}{Johannes Pfeifer}, \bibinfo{person}{Marija Popovi{\'c}},
  \bibinfo{person}{Ciro Potena}, \bibinfo{person}{C{\'e}dric Pradalier},
  \bibinfo{person}{Elisa {Rothacker-Feder}}, \bibinfo{person}{Inkyu Sa},
  \bibinfo{person}{Alexander Schaefer}, \bibinfo{person}{Roland Siegwart},
  \bibinfo{person}{Cyrill Stachniss}, \bibinfo{person}{Achim Walter},
  \bibinfo{person}{Wera Winterhalter}, \bibinfo{person}{Xiaolong Wu}, {and}
  \bibinfo{person}{Juan Nieto}.} \bibinfo{year}{2021}\natexlab{}.
\newblock \showarticletitle{Building an Aerial-Ground Robotics System for
  Precision Farming: An Adaptable Solution}.
\newblock \bibinfo{journal}{\emph{IEEE Robotics \& Automation Magazine}}
  \bibinfo{volume}{28}, \bibinfo{number}{3} (\bibinfo{date}{Sept.}
  \bibinfo{year}{2021}), \bibinfo{pages}{29--49}.
\newblock
\showISSN{1070-9932, 1558-223X}
\urldef\tempurl%
\url{https://doi.org/10.1109/mra.2020.3012492}
\showDOI{\tempurl}


\bibitem[Quinton et~al\mbox{.}(2023)]%
        {quintonMarketApproachesMultiRobot2023a}
\bibfield{author}{\bibinfo{person}{F{\'e}lix Quinton},
  \bibinfo{person}{Christophe Grand}, {and} \bibinfo{person}{Charles Lesire}.}
  \bibinfo{year}{2023}\natexlab{}.
\newblock \showarticletitle{Market {{Approaches}} to the {{Multi-Robot Task
  Allocation Problem}}: A {{Survey}}}.
\newblock \bibinfo{journal}{\emph{Journal of Intelligent \& Robotic Systems}}
  \bibinfo{volume}{107}, \bibinfo{number}{2} (\bibinfo{date}{Feb.}
  \bibinfo{year}{2023}), \bibinfo{pages}{29}.
\newblock
\showISSN{1573-0409}
\urldef\tempurl%
\url{https://doi.org/10.1007/S10846-022-01803-0}
\showDOI{\tempurl}


\bibitem[Schuster et~al\mbox{.}(2020)]%
        {schusterARCHESSpaceAnalogueDemonstration2020}
\bibfield{author}{\bibinfo{person}{Martin~J. Schuster},
  \bibinfo{person}{Marcus~G. M{\"u}ller}, \bibinfo{person}{Sebastian~G.
  Brunner}, \bibinfo{person}{H. Lehner}, \bibinfo{person}{P. Lehner},
  \bibinfo{person}{R. Sakagami}, \bibinfo{person}{A. D{\"o}mel},
  \bibinfo{person}{Lukas Meyer}, \bibinfo{person}{B. Vodermayer},
  \bibinfo{person}{R. Giubilato}, \bibinfo{person}{M. Vayugundla},
  \bibinfo{person}{Josef Reill}, \bibinfo{person}{Florian Steidle},
  \bibinfo{person}{Ingo {von Bargen}}, \bibinfo{person}{K. Bussmann},
  \bibinfo{person}{Rico Belder}, \bibinfo{person}{P. Lutz}, \bibinfo{person}{W.
  St{\"u}rzl}, \bibinfo{person}{Michal S.}, \bibinfo{person}{Moritz Maier},
  \bibinfo{person}{S. Stoneman}, \bibinfo{person}{Andre~F. Prince},
  \bibinfo{person}{B. Rebele}, \bibinfo{person}{M. Durner}, \bibinfo{person}{E.
  Staudinger}, \bibinfo{person}{Siwei Zhang}, \bibinfo{person}{Robert P.},
  \bibinfo{person}{Esther Bischoff}, \bibinfo{person}{C. Braun},
  \bibinfo{person}{Susanne Schr{\"o}der}, \bibinfo{person}{E. Dietz},
  \bibinfo{person}{S. Frohmann}, \bibinfo{person}{Anko B{\"o}rner},
  \bibinfo{person}{H. H{\"u}bers}, \bibinfo{person}{B. Foing},
  \bibinfo{person}{R. Triebel}, \bibinfo{person}{Alin~O. Albu}, {and}
  \bibinfo{person}{A. Wedler}.} \bibinfo{year}{2020}\natexlab{}.
\newblock \showarticletitle{The {{ARCHES Space-Analogue Demonstration
  Mission}}: {{Towards Heterogeneous Teams}} of {{Autonomous Robots}} for
  {{Collaborative Scientific Sampling}} in {{Planetary Exploration}}}.
\newblock \bibinfo{journal}{\emph{IEEE Robotics and Automation Letters}}
  \bibinfo{volume}{5}, \bibinfo{number}{4} (\bibinfo{date}{Oct.}
  \bibinfo{year}{2020}), \bibinfo{pages}{5315--5322}.
\newblock
\showISSN{2377-3766}
\urldef\tempurl%
\url{https://doi.org/10.1109/LRA.2020.3007468}
\showDOI{\tempurl}


\bibitem[Sedlmeir et~al\mbox{.}(2020)]%
        {sedlmeir2020energy}
\bibfield{author}{\bibinfo{person}{Johannes Sedlmeir},
  \bibinfo{person}{Hans~Ulrich Buhl}, \bibinfo{person}{Gilbert Fridgen}, {and}
  \bibinfo{person}{Robert Keller}.} \bibinfo{year}{2020}\natexlab{}.
\newblock \showarticletitle{The Energy Consumption of Blockchain Technology:
  Beyond Myth}.
\newblock \bibinfo{journal}{\emph{Business \& Inf. Sys. Eng.}}
  \bibinfo{volume}{62}, \bibinfo{number}{6} (\bibinfo{year}{2020}),
  \bibinfo{pages}{599--608}.
\newblock
\urldef\tempurl%
\url{https://doi.org/10.1007/s12599-020-00656-x}
\showDOI{\tempurl}


\bibitem[Sedlmeir et~al\mbox{.}(2022)]%
        {sedlmeir2022transparency}
\bibfield{author}{\bibinfo{person}{Johannes Sedlmeir},
  \bibinfo{person}{Jonathan Lautenschlager}, \bibinfo{person}{Gilbert Fridgen},
  {and} \bibinfo{person}{Nils Urbach}.} \bibinfo{year}{2022}\natexlab{}.
\newblock \showarticletitle{The Transparency Challenge of Blockchain in
  Organizations}.
\newblock \bibinfo{journal}{\emph{Electronic Markets}}  \bibinfo{volume}{32}
  (\bibinfo{year}{2022}), \bibinfo{pages}{1779---1794}.
\newblock
\urldef\tempurl%
\url{https://doi.org/10.1007/s12525-022-00536-0}
\showDOI{\tempurl}


\bibitem[Strobel et~al\mbox{.}(2018)]%
        {strobelManagingByzantineRobots2018}
\bibfield{author}{\bibinfo{person}{Volker Strobel}, \bibinfo{person}{Eduardo
  Castell{\'o}~Ferrer}, {and} \bibinfo{person}{Marco Dorigo}.}
  \bibinfo{year}{2018}\natexlab{}.
\newblock \showarticletitle{Managing {{Byzantine}} Robots via Blockchain
  Technology in a Swarm Robotics Collective Decision Making Scenario}. In
  \bibinfo{booktitle}{\emph{Proc. 17th {{Int}}. {{Conf}}. {{Auton}}. {{Agents
  MultiAgent Syst}}.}} \emph{(\bibinfo{series}{{{AAMAS}} '18})}.
  \bibinfo{publisher}{{International Foundation for Autonomous Agents and
  Multiagent Systems}}, \bibinfo{address}{{Stockholm, Sweden}},
  \bibinfo{pages}{541--549}.
\newblock
\urldef\tempurl%
\url{https://doi.org/10.5555/3237383.3237464}
\showDOI{\tempurl}


\bibitem[Triantafyllou et~al\mbox{.}(2021)]%
        {triantafyllouMethodologyApproachingIntegration2021a}
\bibfield{author}{\bibinfo{person}{Pavlos Triantafyllou},
  \bibinfo{person}{Rafael Afonso~Rodrigues}, \bibinfo{person}{Sirapoab
  Chaikunsaeng}, \bibinfo{person}{Diogo Almeida}, \bibinfo{person}{Graham
  Deacon}, \bibinfo{person}{Jelizaveta Konstantinova}, {and}
  \bibinfo{person}{Giuseppe Cotugno}.} \bibinfo{year}{2021}\natexlab{}.
\newblock \showarticletitle{A Methodology for Approaching the Integration of
  Complex Robotics Systems: {{Illustration}} through a Bimanual Manipulation
  Case Study}.
\newblock \bibinfo{journal}{\emph{IEEE Robotics \& Automation Magazine}}
  \bibinfo{volume}{28} (\bibinfo{year}{2021}), \bibinfo{pages}{88}.
\newblock
\showISSN{1070-9932, 1558-223X}
\urldef\tempurl%
\url{https://doi.org/10.1109/mra.2021.3064759}
\showDOI{\tempurl}


\bibitem[Weinzierl and Sarang(2021)]%
        {weinzierlCommercialSpaceAge2021}
\bibfield{author}{\bibinfo{person}{Matthew Weinzierl} {and}
  \bibinfo{person}{Mehak Sarang}.} \bibinfo{year}{2021}\natexlab{}.
\newblock \showarticletitle{The Commercial Space Age Is Here}.
\newblock \bibinfo{journal}{\emph{Harvard Busi. Rev}}  \bibinfo{volume}{.}
  (\bibinfo{year}{2021}), \bibinfo{pages}{1}.
\newblock
\showISSN{0017-8012}
\urldef\tempurl%
\url{hbr.org/2021/02/the-commercial-space-age-is-here}
\showURL{%
\tempurl}


\bibitem[Zheng et~al\mbox{.}(2020)]%
        {zhengAIEconomistImproving2020}
\bibfield{author}{\bibinfo{person}{Stephan Zheng}, \bibinfo{person}{Alexander
  Trott}, \bibinfo{person}{Sunil Srinivasa}, \bibinfo{person}{Nikhil Naik},
  \bibinfo{person}{Melvin Gruesbeck}, \bibinfo{person}{David~C. Parkes}, {and}
  \bibinfo{person}{Richard Socher}.} \bibinfo{year}{2020}\natexlab{}.
\newblock \bibinfo{title}{The {{AI}} Economist: Improving Equality and
  Productivity with {{AI-driven}} Tax Policies}.
\newblock
\newblock
\urldef\tempurl%
\url{https://doi.org/10.48550/arXiv.2004.13332}
\showDOI{\tempurl}
\showeprint[arxiv]{2004.13332}


\bibitem[Zlot et~al\mbox{.}(2002)]%
        {zlotMultirobotExplorationControlled2002a}
\bibfield{author}{\bibinfo{person}{Robert Zlot}, \bibinfo{person}{Anthony
  Stentz}, \bibinfo{person}{M.Bernardine Dias}, {and} \bibinfo{person}{Scott
  Thayer}.} \bibinfo{year}{2002}\natexlab{}.
\newblock \showarticletitle{Multi-Robot Exploration Controlled by a Market
  Economy}. In \bibinfo{booktitle}{\emph{Proceedings 2002 {{IEEE International
  Conference}} on {{Robotics}} and {{Automation}} ({{Cat}}.
  {{No}}.{{02CH37292}})}}, Vol.~\bibinfo{volume}{3}.
  \bibinfo{publisher}{{IEEE}}, \bibinfo{address}{{DC, USA}},
  \bibinfo{pages}{3016--3023}.
\newblock
\showISBNx{978-0-7803-7272-6}
\urldef\tempurl%
\url{https://doi.org/10.1109/ROBOT.2002.1013690}
\showDOI{\tempurl}


\end{thebibliography}

\end{document}